%% file: main.tex
\documentclass[10pt,twocolumn,letterpaper]{article}

\usepackage{cvpr}              

\input{preamble}

\definecolor{cvprblue}{rgb}{0.21,0.49,0.74}
\usepackage[pagebackref,breaklinks,colorlinks,allcolors=cvprblue]{hyperref}

\def\paperID{16} 
\def\confName{CVPR}
\def\confYear{2026}

\title{Consistent Yet Wrong: Evidence Insensitivity in\\
Spatial Vision-Language Models}

\author{S Divakar Bhat \quad Toshihiko Yamasaki \\
{ The University of Tokyo, Japan} \\
{\tt\small \{bhat, yamasaki\}@cvm.t.u-tokyo.ac.jp}
}

\begin{document}
\twocolumn[{
\maketitle
\vspace{-0.15cm}
\begin{center}
    \includegraphics[width=\linewidth,trim={0cm, 0cm, 0cm, 0cm}, clip]{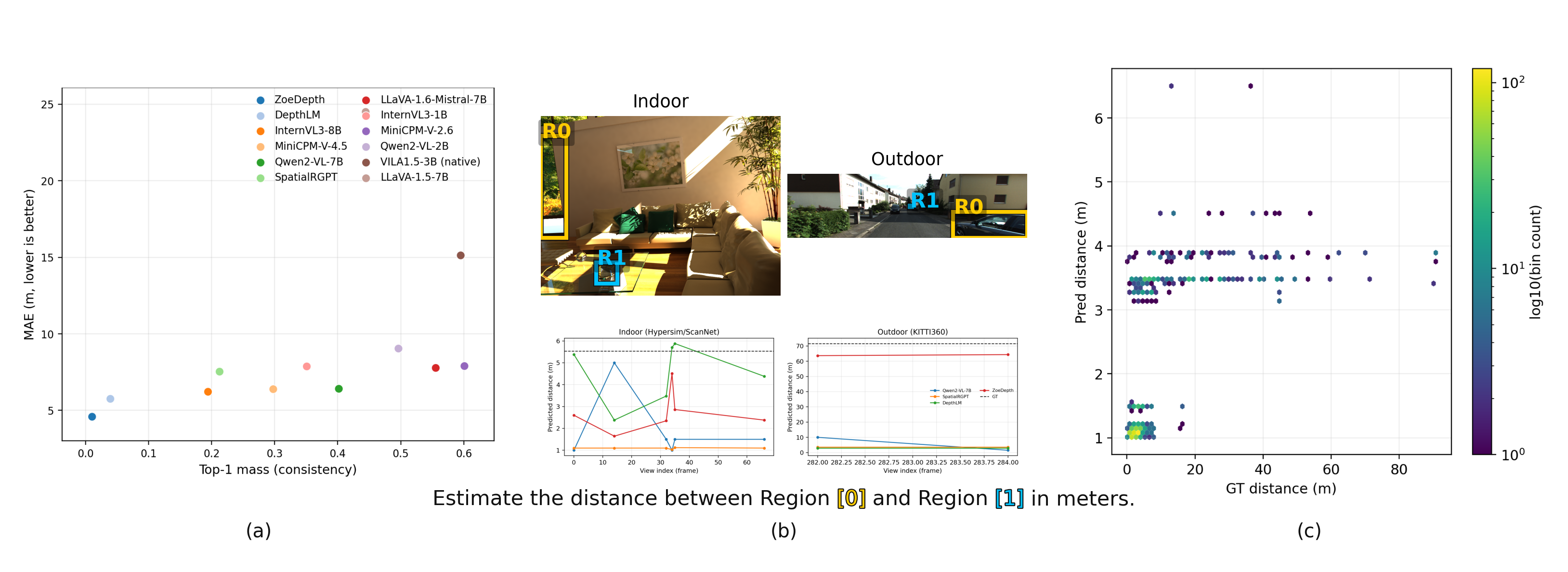}
    \captionof{figure}{\emph{View consistency can mislead}: geometry baselines such as ZoeDepth~\cite{bhat2023zoedepth} track visual evidence and achieve lower error, while typical and spatial VLMs are often consistent yet wrong. \textbf{[Left]} The collapse quadrant (Consistency vs.\ Accuracy) highlights models that are highly consistent but inaccurate. \textbf{[Centre]} Same-pair traces show predictions that remain fixed across views despite changing evidence. \textbf{[Right]} A biased GT--prediction density for SpatialRGPT~\cite{cheng2024spatialrgpt} exposes prior-driven errors. ViewDiag and our diagnostics test evidence sensitivity beyond accuracy alone.}
    \label{fig:teaser}
\end{center}
\vspace{0.1cm}
}]
\input{sec/0_abstract}
\input{sec/1_intro}
\input{sec/2_related}
\input{sec/4_protocol}
\input{sec/3_formulation}
\input{sec/5_metrics}
\input{sec/5_experiments}
\input{sec/6_results}
\input{sec/7_analysis}

\input{sec/8_limitations}
\input{sec/9_conclusion}
{
    \small
    \clearpage
    \bibliographystyle{ieeenat_fullname}
    \bibliography{main}
}


\end{document}

%% file: preamble.tex
\usepackage{caption}
\usepackage{algorithm}
\usepackage{algpseudocode}
\usepackage{balance}
\usepackage{float}
\usepackage{etoolbox}
\apptocmd{\thebibliography}{%
  \setlength{\parskip}{0pt}%
  \setlength{\itemsep}{0pt}%
}{}{}
\usepackage[section]{placeins}

\usepackage{cuted}

\usepackage{cuted}


%% file: sec/0_abstract.tex
\begin{abstract}
Spatial reasoning is fundamental to robotics, autonomy, and embodied AI, yet modern vision-language models (VLMs) remain unreliable on metric distance queries. A common assumption is that consistent predictions across viewpoints reflect geometric grounding. We test this assumption and find the opposite: leading VLMs often produce view-invariant and consistent  answers even when those answers are incorrect, indicating weak coupling between predictions and viewpoint-specific visual evidence.

We introduce \textbf{ViewDiag}, a controlled multi-view evaluation protocol built from Hypersim, ScanNet, and KITTI360, comprising 176 object-pair tracks across 80 scenes with 2--10 views per track. The protocol evaluates models along three axes: metric accuracy, distributional concentration, and internal collapse, the last of which is assessed using a latent feature probe. Across diverse models, we observe a consistent pattern of high prediction stability paired with substantial error, clustering in a regime characterized by strong consistency but low accuracy.

\noindent These results challenge the common use of cross-view consistency as a proxy for geometric understanding. Instead, we show that stable predictions may reflect prior-driven collapse rather than evidence-sensitive reasoning. ViewDiag provides a controlled benchmark and diagnostic framework for evaluating whether spatial VLMs are not only accurate, but also meaningfully coupled to visual evidence.
\end{abstract}

%% file: sec/1_intro.tex
\section{Introduction}
\label{sec:intro}

Spatial reasoning is a core competency for robotics, autonomous driving, embodied agents, and navigation systems. Many deployments require reliable metric judgments: estimating the distance to a curb, measuring the separation between objects, or assessing the length of a corridor. VLMs have recently demonstrated strong general reasoning and improved spatial reasoning capabilities, yet their behavior on metric distance queries remains brittle, particularly under viewpoint changes and long-range geometry~\cite{wu2025indoor}.

A common assumption in the literature is that prediction consistency across viewpoints implies geometric grounding~\cite{feng2025seeing}. For a given question, if a model produces the same distance estimate from different views, the prediction is often seen as evidence of stable scene understanding. This assumption is appealing because it offers a simple validation signal without requiring explicit geometry pipelines or additional sensors: stable outputs appear to reflect viewpoint-invariant reasoning. This raises a key question: \emph{Does cross-view consistency reflect genuine geometric grounding, or can it arise from prior-driven prediction collapse?} We do not argue that consistency is undesirable. Rather, we show that consistency is informative only when paired with accuracy and evidence dependence.

\begin{table}[t]
\centering
\small
\begin{tabular}{l p{0.62\linewidth}}
\toprule
Field & ViewDiag \\
\midrule
Inputs & RGB image, bounding-box pairs, prompt, camera metadata \\
Outputs & Metric distance (m) + identifiers for cross-view linking \\
Datasets & Hypersim, ScanNet, KITTI360 (indoor/outdoor, synthetic/real) \\
Tracks / queries & 176 region pair tracks, 1308 queries \\
Views per track & 2--10 (mean 7.43) \\
Model families & VLMs, geometry-only depth baselines, and hybrids \\
\bottomrule
\end{tabular}
\caption{Benchmark utility summary for ViewDiag data.}
\label{tab:benchmark_utility}
\end{table}
We test this assumption directly by examining whether spatial VLM predictions actually track visual evidence across viewpoints or instead default to learned priors. We refer to the latter behavior as \emph{evidence insensitivity}: predictions remain largely unchanged under controlled viewpoint variations even when the visual evidence differs. Conversely, we call a model evidence-sensitive if its predictions and internal states respond appropriately to viewpoint specific visual cues while preserving the invariant metric quantity. This distinction matters because stability may arise from two fundamentally different mechanisms: genuine geometric reasoning or reliance on learned priors.

Across several leading VLM families, we observe consistent predictions across views even when those predictions are incorrect. This \emph{consistent-yet-wrong} behavior is the defining symptom of evidence insensitivity. Rather than reflecting geometric reasoning, the stability often arises from a small set of prior-dominated outputs that are weakly coupled to the visual evidence in the image. In other words, a model can appear robust while failing to meaningfully react to changes in viewpoint.
\begin{figure}[t]
    \centering
    \includegraphics[width=\linewidth]{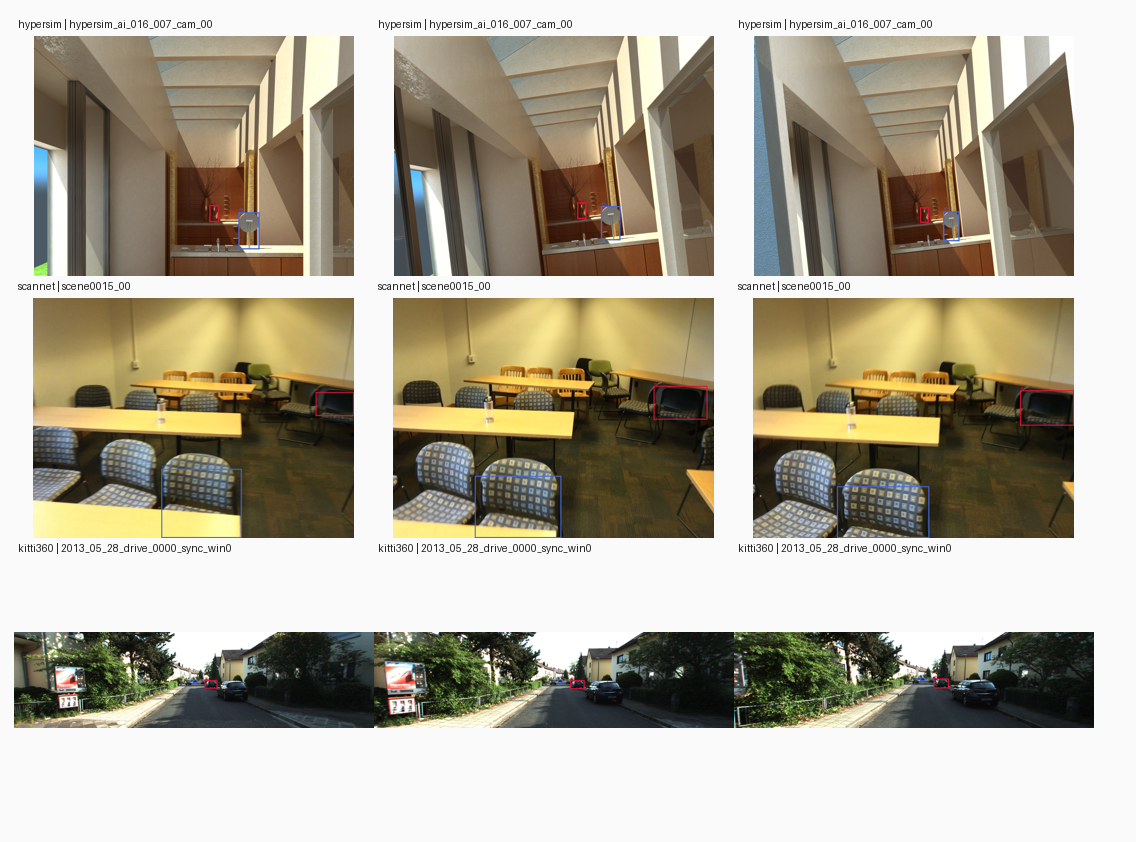}
    \caption{ViewDiag samples with annotated regions ([R0]/[R1]): three examples per dataset (Hypersim, ScanNet, KITTI360). The same region pair is tracked across multiple viewpoints to test evidence sensitivity.}
    \label{fig:dataset_panel}
\end{figure}
To analyze this failure mode, we introduce \textbf{ViewDiag}, a controlled multi-view evaluation protocol and diagnostic toolkit for spatial reasoning in VLMs. As shown in Table \ref{tab:benchmark_utility}, ViewDiag constructs multi-view region pairs with fixed queries across views, enabling direct measurement of evidence sensitivity beyond single-view accuracy. We evaluate predictions along three axes: metric accuracy, distributional consistency, and internal collapse. A latent-feature probe compares hidden states across same-output views to test whether invariance reflects stable representations or unstable decision mappings. Our experiments show that many models cluster in a collapse regime marked by high consistency but low accuracy, indicating that stability alone is not a reliable signal of geometric understanding.

\noindent\textbf{Practical implications for multimodal systems.}
In robotics and autonomous driving, cross-view or cross-sensor consistency is often treated as a proxy for robustness. Our results caution otherwise: stable but incorrect outputs may be more dangerous than uncertain ones. ViewDiag provides a diagnostic framework for model selection and validation, highlights where geometry-only or hybrid baselines remain important, and identifies long-range outdoor scenes as a regime where evidence sensitivity degrades.

\begin{figure*}[!t]
	\centering
	\begin{minipage}{0.32\linewidth}
		\centering
		\includegraphics[width=\linewidth, trim={0cm, 0cm, 0cm, 0.73cm}, clip]{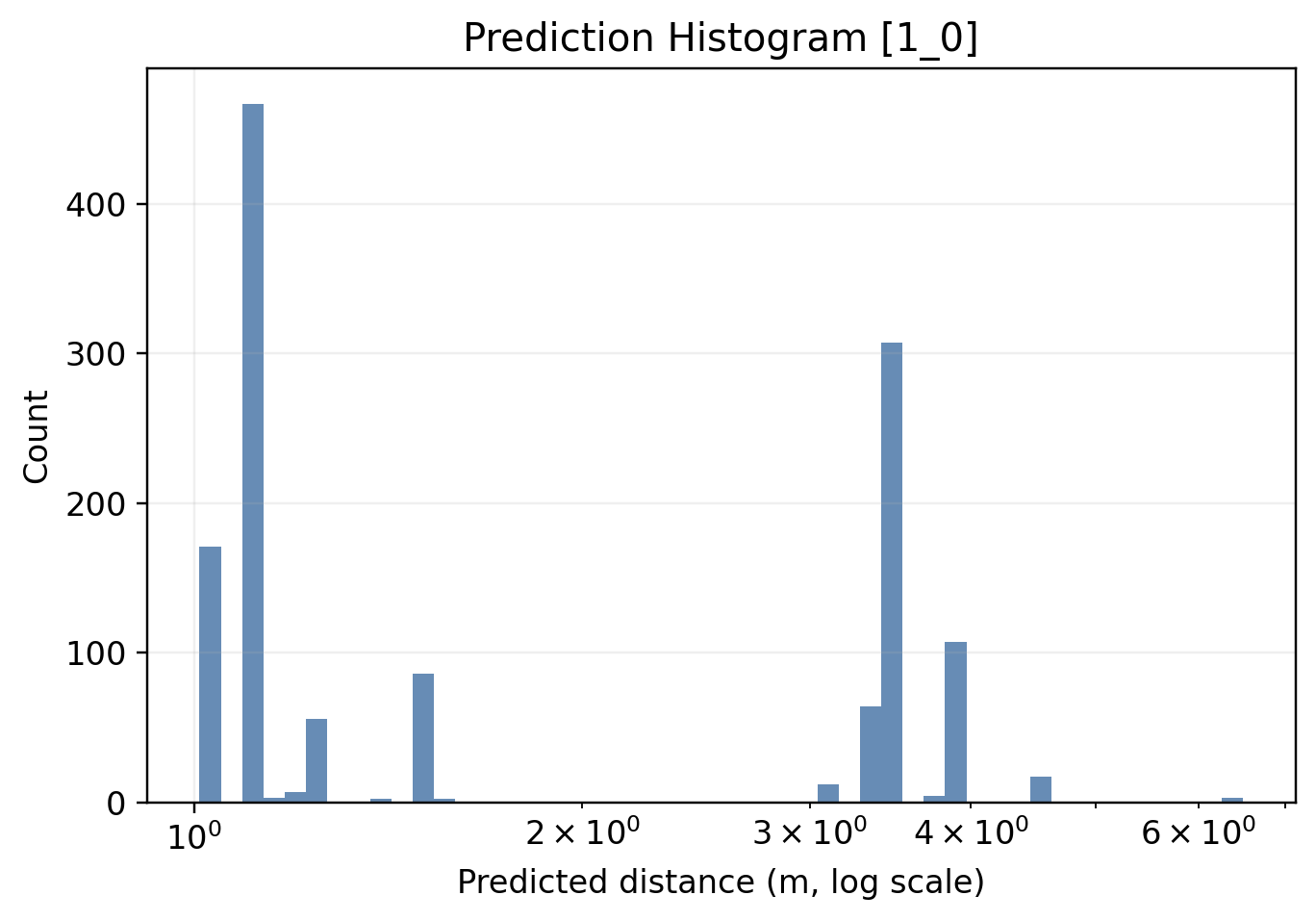}
		\small SpatialRGPT
	\end{minipage}
	\begin{minipage}{0.32\linewidth}
		\centering
		\includegraphics[width=\linewidth, trim={0cm, 0cm, 0cm, 0.73cm}, clip]{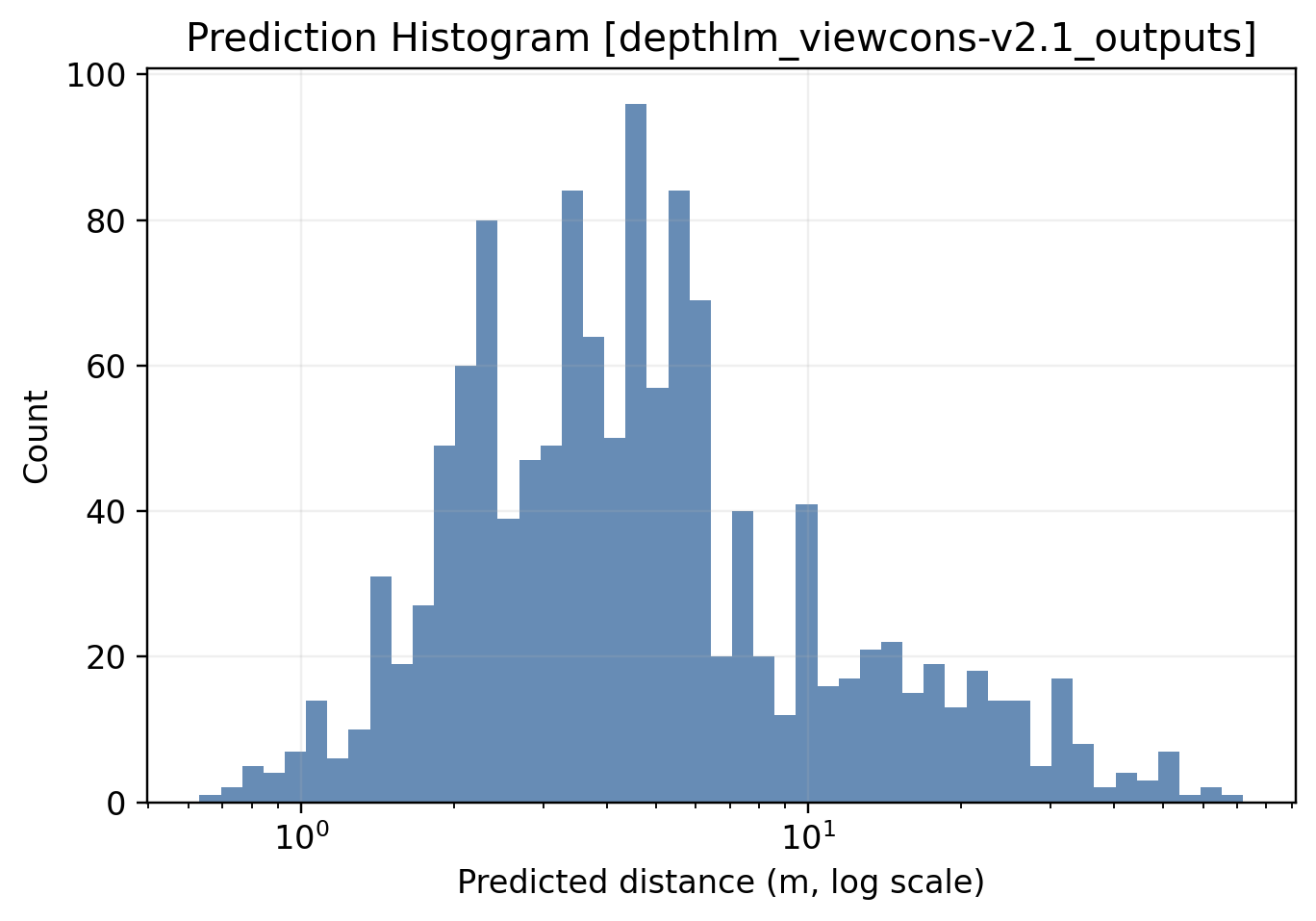}
		\small DepthLM
	\end{minipage}
	\begin{minipage}{0.32\linewidth}
		\centering
		\includegraphics[width=\linewidth,trim={0cm, 0cm, 0cm, 0.73cm}, clip]{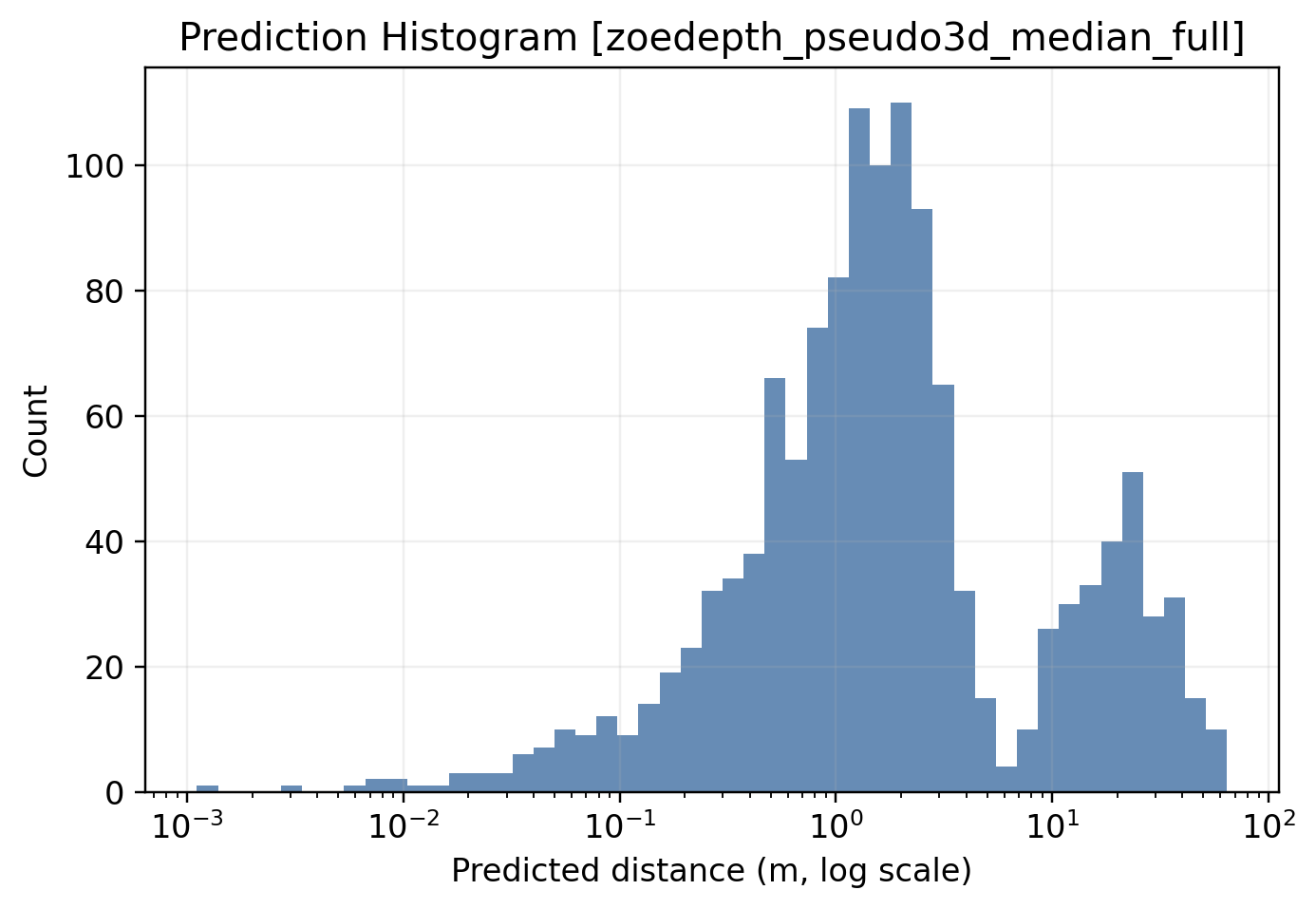}
		\small ZoeDepth
	\end{minipage}
	\caption{Prediction histograms (log bins) for SpatialRGPT, DepthLM, and ZoeDepth. Geometry baselines exhibit broader output distributions than spatial VLMs, suggesting better sensitivity to evidence in geometrically grounded models. }
	\label{fig:pred_hist_models}
\end{figure*}

\noindent\textbf{Contributions.}
\begin{itemize}
    \item A controlled multi-view evaluation protocol for spatial distance queries using real multi-view region pairs.
    \item Empirical evidence that view-consistent predictions can remain systematically incorrect, indicating evidence-insensitive behavior in spatial VLMs.
    \item A benchmark and evaluation framework for analyzing evidence sensitivity beyond single-view accuracy that jointly measures accuracy, output concentration, and internal collapse to distinguish evidence sensitive reasoning from prior-driven stability.
\end{itemize}
Beyond introducing a diagnostic framework, our study highlights an underexplored failure mode in spatial vision--language models: consistent yet incorrect predictions under viewpoint change. This exposes a fundamental gap between apparent robustness and genuine evidence sensitivity.

%% file: sec/2_related.tex
\section{Related Work}
\label{sec:related}
\begin{figure}[t]
    \centering
    \includegraphics[width=\linewidth]{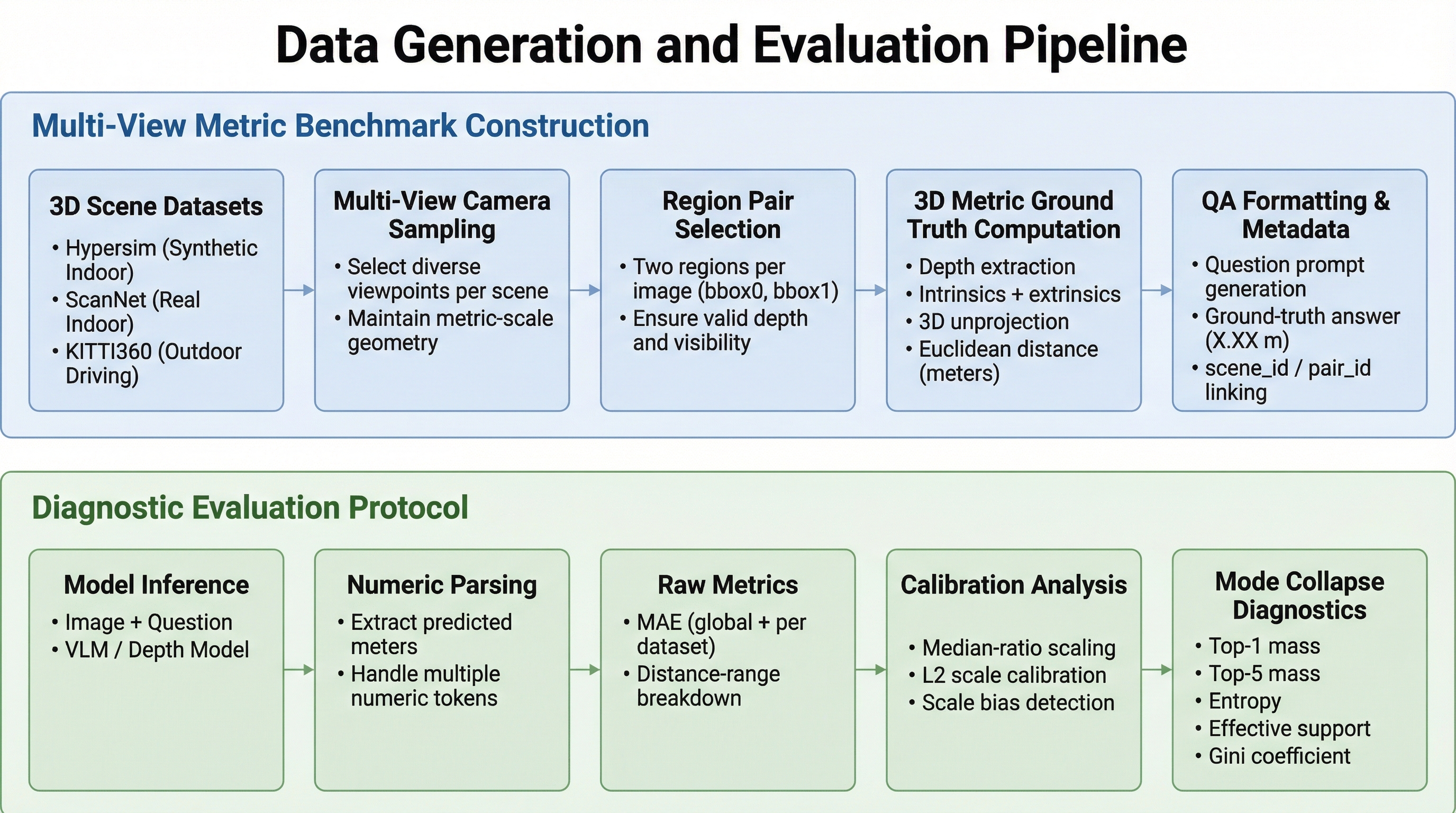}
    \caption{Controlled multi-view protocol. A fixed region pair is queried across views, then evaluated with accuracy, consistency, and internal-collapse diagnostics.}
    \label{fig:pipeline}
\end{figure}
\noindent\textbf{Vision-language models and spatial reasoning.}
Large vision--language models (VLMs) have substantially improved multimodal reasoning, visual question answering, and instruction following. Foundational models such as CLIP~\cite{radford2021clip}, BLIP~\cite{li2022blip}, BLIP-2~\cite{li2023blip2}, and Flamingo~\cite{alayrac2022flamingo} established scalable vision--language alignment, while instruction-tuned systems including LLaVA~\cite{liu2023llava}, VILA~\cite{lin2024vila}, Qwen2-VL~\cite{qwen2vl2024}, InternVL~\cite{chen2024internvl}, and MiniCPM-V~\cite{minicpmv2024} further improved open-ended multimodal reasoning. However, strong performance on broad benchmarks does not necessarily imply reliable metric or geometry-sensitive reasoning. Specialized efforts such as SpatialVLM~\cite{chen2024spatialvlm}, SpatialRGPT~\cite{cheng2024spatialrgpt}, and 3D-LLM~\cite{hong2023d3llm} incorporate spatial supervision or 3D structure, but do not directly test whether predictions remain sensitive to viewpoint-specific evidence under fixed cross-view queries.
\begin{table}[t]
\centering
\small
\begin{tabular}{lc p{0.32\linewidth}}
\toprule
Dataset & Samples & Notes \\
\midrule
Hypersim & 941 & indoor, synthetic \\
ScanNet & 72 & indoor, real \\
KITTI360 & 295 & outdoor, real \\
\bottomrule
\end{tabular}
\caption{ViewDiag composition by source dataset. Samples are multi-view region pairs with metric distance supervision.}
\label{tab:dataset_stats}
\end{table}
\noindent\textbf{Benchmarks for spatial reasoning.}
Spatial reasoning has long been studied through diagnostic benchmarks. CLEVR~\cite{johnson2017clevr} provides minimally biased visual reasoning tasks, while GQA~\cite{hudson2019gqa} extends compositional reasoning to real images and introduces metrics such as consistency and grounding. More recent benchmarks accompany specialized spatial models, including those introduced with SpatialRGPT~\cite{cheng2024spatialrgpt} and SpatialVLM~\cite{chen2024spatialvlm}. These benchmarks measure spatial task accuracy, but generally do not isolate whether stable predictions under viewpoint change are evidence-driven or prior-dominated.

\noindent\textbf{Multi-view reasoning and viewpoint robustness.}
Recent work studies whether multimodal models can reason consistently across viewpoints. All-Angles Bench reports that current MLLMs remain weak at cross-view correspondence, relative distance, relative direction, and camera-pose reasoning~\cite{yeh2025allangles}. In contrast to such broad multi-view QA settings, our focus is narrower: we test whether a fixed region-pair metric query remains both accurate and evidence-sensitive across views.

\noindent\textbf{Depth and metric geometry.}
Metric spatial reasoning is closely related to monocular depth estimation. Geometry-centered methods provide strong baselines because they are optimized for scene structure rather than open-ended language generation. MegaDepth~\cite{li2018megadepth}, Monodepth2~\cite{godard2019monodepth2}, and DPT~\cite{ranftl2021dpt} established influential paradigms for single-image depth prediction, while Depth Anything~\cite{yang2024depthanything} and ZoeDepth~\cite{bhat2023zoedepth} improve robustness and cross-domain generalization. DepthLM~\cite{cai2025depthlm} is particularly relevant because it adapts VLMs for metric depth prediction. These methods provide important comparators by separating metric accuracy from evidence sensitivity.

\noindent\textbf{Robustness and shortcut learning.}
Neural models often exploit narrow statistical regularities rather than intended signals~\cite{geirhos2020shortcut}. In spatial VLMs, this can appear as view-stable predictions with incorrect geometry. We therefore pair accuracy with concentration metrics and internal probes to interpret prediction stability.

\noindent\textbf{Positioning.}
Prior work largely evaluates spatial reasoning by accuracy. We instead study \emph{evidence dependence}: whether consistent predictions reflect geometric grounding or collapse to prior-driven outputs. Figure~\ref{fig:pipeline} shows our controlled multi-view protocol.

%% file: sec/4_protocol.tex
\section{Controlled Multi-View Evaluation Protocol}
\label{sec:protocol}

\noindent\textbf{Dataset goal and sources.}
ViewDiag is a geometry-grounded diagnostic benchmark for metric distance reasoning under multi-view conditions. Each example queries the distance between two image regions and derives ground truth directly from scene geometry rather than human annotation. The dataset is constructed from three complementary sources: Hypersim~\cite{roberts2021hypersim} (synthetic indoor), ScanNet~\cite{dai2017scannet} (real indoor), and KITTI360~\cite{liao2021kitti360} (outdoor driving) as shown in Table \ref{tab:dataset_stats}. This combination provides diversity across indoor and outdoor environments, synthetic and real imagery, and both short- and long-range distances.

\noindent\textbf{Benchmark construction and validity checks.}
We apply a compact set of gates to ensure regions are visible, depth-supported, and stable across views. Concretely, we require a minimum instance area (1024 px for Hypersim/ScanNet, 128 px for KITTI360) and valid depth support (at least 60 pixels for indoor data). Depth is filtered around the median with a tight band ($\lvert d - \mathrm{median}(d) \rvert \leq 0.1$ m), and we reject depth-inconsistent regions ($\sigma/\mu > 0.40$). We also filter region pairs with excessive overlap or implausible geometry (e.g., IoU $\leq 0.20$ for indoor pairs) and enforce multi-view stability by bounding center drift and area change across views (center-std $\leq 0.20$ indoor, $\leq 0.22$ KITTI360; area-ratio caps 4.0/4.5). Minimum valid views per region pair are 4 (Hypersim), 3 (ScanNet), and 2 (KITTI360).

\noindent\textbf{Multi-view sampling and region pairs.}
For each scene, we select multiple camera views in which the same region pair remains visible while preserving camera intrinsics, extrinsics, and depth support. Regions are required to be non-degenerate and span diverse depths and spatial layouts. Each sample consists of one image, two regions, and a metric distance query, while multiple views of the same pair enable evidence-sensitivity analysis. We use ``region pair'' for the two queried boxes in one image, and a ``region-pair track'' for that same pair across multiple views. In total, ViewDiag contains 1308 queries across 176 region pair tracks and 80 scenes, with 2--10 views per region pair (mean 7.43). Figure~\ref{fig:dataset_panel} shows example samples, and illustrates multi-view pairs.

\noindent\textbf{Design rationale.}
We target 2--10 views per region pair to separate evidence sensitivity from single-view accuracy: a grounded model should remain accurate while its internal evidence changes across viewpoints. The dataset mix provides both indoor and outdoor coverage, spanning short indoor distances and longer outdoor driving scenes where prior-dominated collapse becomes more visible. Region-pair tracks are sampled to cover diverse spatial layouts while avoiding near-duplicate views.

\noindent\textbf{Metric ground truth and QA formatting.}
For each region pair, we compute robust depth statistics, unproject them to 3D using dataset camera intrinsics, and measure Euclidean distance in meters. This preserves metric scale while reducing annotation noise. We serialize each sample as a VQA-style prompt, e.g., ``Estimate the distance between Region[0] and Region[1] in meters,'' with metadata for question, scene, pair, source dataset, file path, and bounding boxes to support dataset analysis and cross-view linking.

\noindent\textbf{Model input alignment.}
We convert each sample into the required model format: one RGB image path, two bounding boxes, and a standardized prompt. SpatialRGPT receives region masks directly via $<$mask$>$ tokens, while standard VLMs (LLaVA, Qwen, InternVL, MiniCPM, VILA) receive an overlay image with colored boxes labeled 0/1 and prompts referring to Region[0]/Region[1]. 

%% file: sec/3_formulation.tex
\section{Problem Formulation}
\label{sec:formulation}

Let $I_v$ be an image of a fixed scene observed from viewpoint $v$, and let $Q$ be a spatial query asking for the metric distance between two regions (Region [0] and Region [1]). A model $f$ maps the image and query to a predicted distance:
\begin{equation}
    y_v = f(I_v, Q).
\end{equation}
Each viewpoint $v$ corresponds to a distinct camera pose observing the same region pair.

\begin{figure*}[!t]
	\centering
	\begin{minipage}{0.32\linewidth}
		\centering
		\includegraphics[width=\linewidth, trim={0cm, 0cm, 0cm, 0.8cm}, clip]{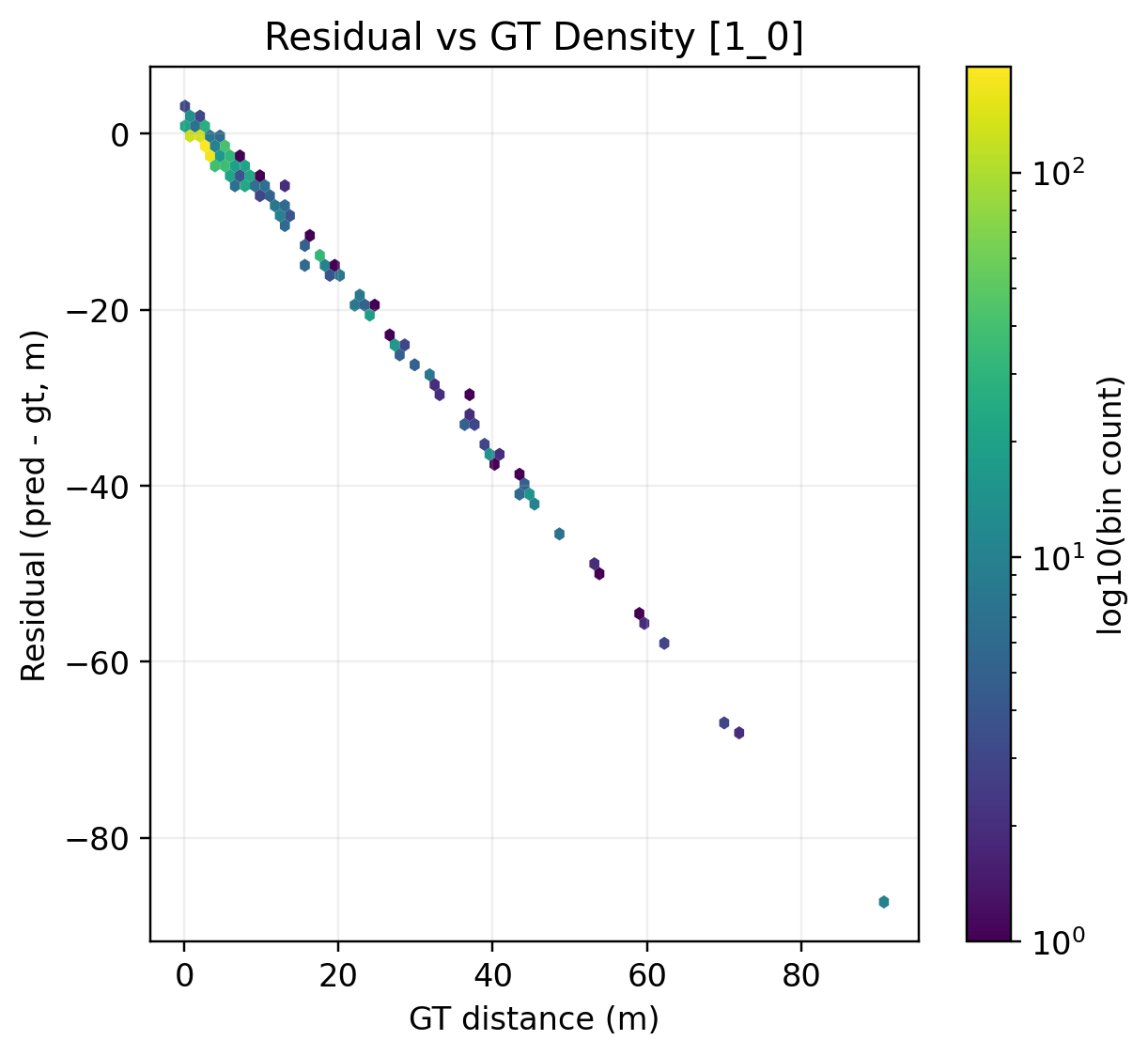}
		\small SpatialRGPT
	\end{minipage}
	\begin{minipage}{0.32\linewidth}
		\centering
		\includegraphics[width=\linewidth, trim={0cm, 0cm, 0cm, 0.8cm}, clip]{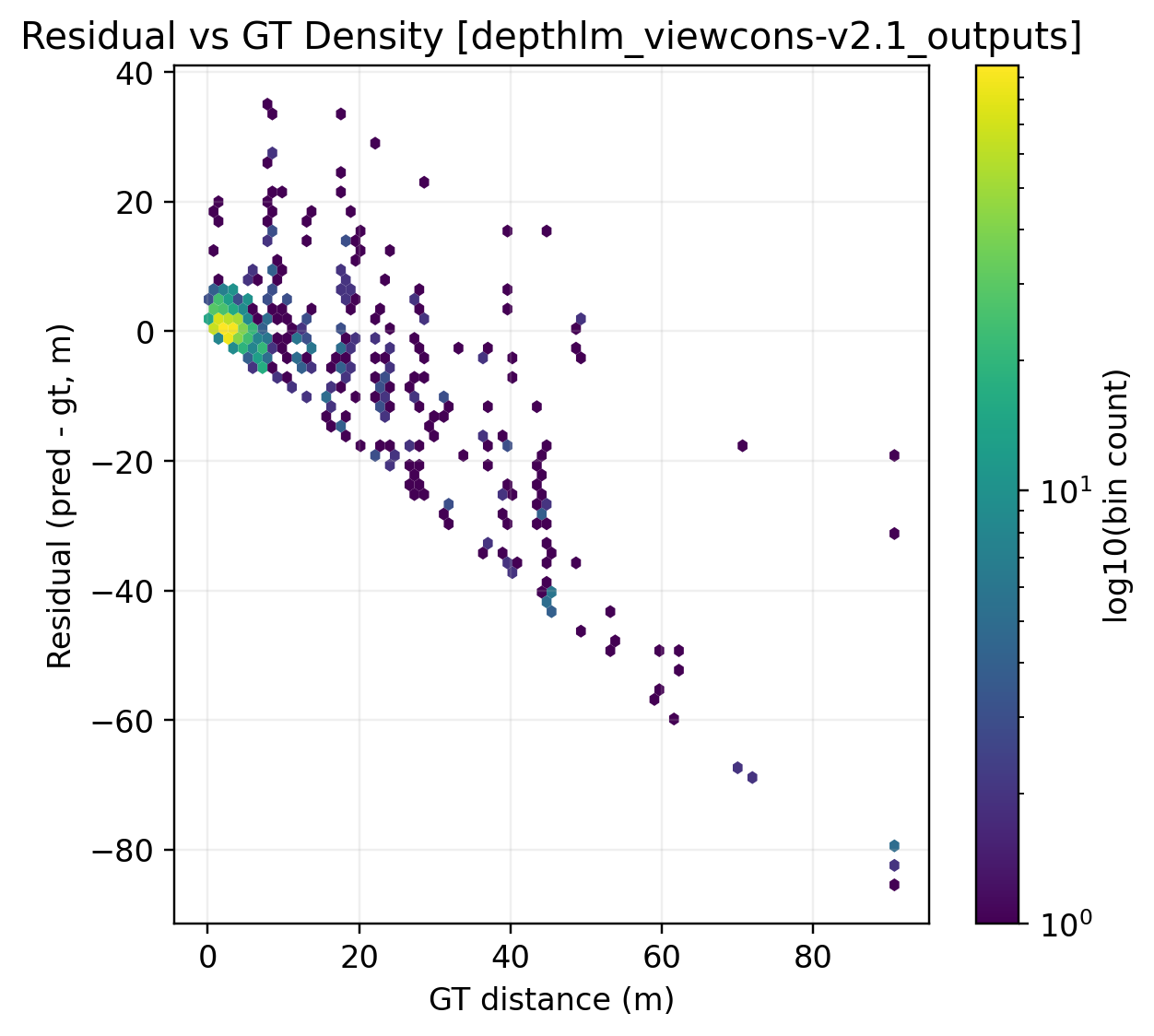}
		\small DepthLM
	\end{minipage}
	\begin{minipage}{0.32\linewidth}
		\centering
		\includegraphics[width=\linewidth, trim={0cm, 0cm, 0cm, 0.8cm}, clip]{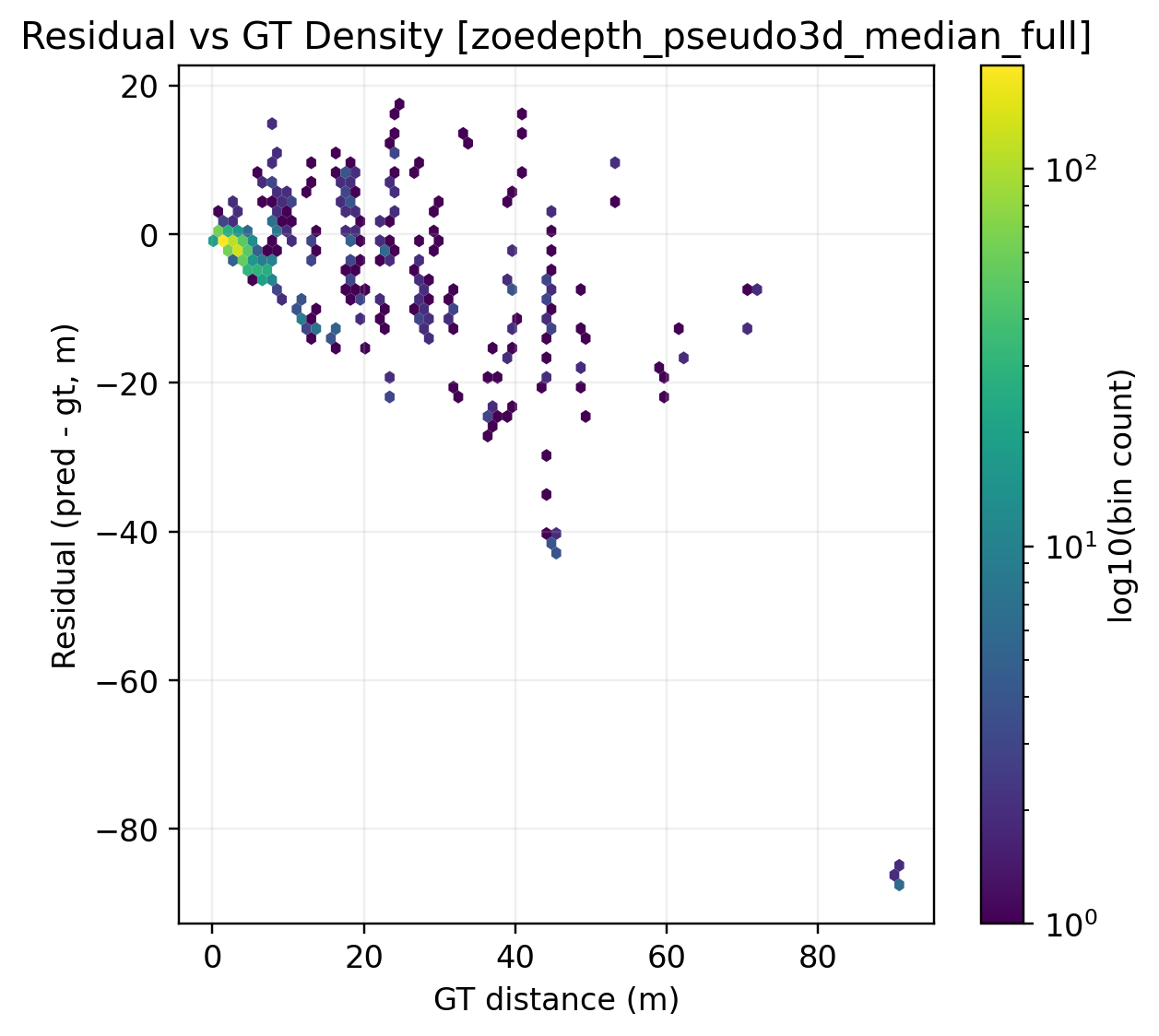}
		\small ZoeDepth
	\end{minipage}
	\caption{Residual vs GT (log bin count) for SpatialRGPT, DepthLM, and ZoeDepth, contrasting spatial VLMs and geometry-based baselines.}
	\label{fig:residual_depth_baselines}
\end{figure*}
\noindent\textbf{View consistency and evidence sensitivity.}
For a fixed region pair, the true metric distance remains invariant across viewpoints. A geometrically grounded model should therefore produce consistent predictions across views when it is correct. The key question is whether such consistency reflects sensitivity to visual evidence or reliance on learned priors that ignore geometric variation.

We define \emph{evidence sensitivity} as meaningful internal variation across in latent representations induced by viewpoint changes, even when the correct output remains stable. Let $h_v$ denote a pooled latent representation for view $v$, and let $\Delta(h_{v_1},h_{v_2})$ measure hidden-state change (e.g., cosine distance or L2 difference). Our diagnostics evaluate pairs of views that produce the same output and examine whether internal signals differ. When predictions remain stable but incorrect while internal representations change, we interpret this behavior as \emph{evidence-insensitive collapse}.

\noindent\textbf{Collapse regime.}
We characterize this phenomenon along two axes: \emph{accuracy} (distance to ground truth) and \emph{consistency} (prediction concentration and cross-view stability). Models that are both inaccurate and highly consistent occupy a collapse regime, where stability arises from prior-driven outputs rather than evidence-sensitive reasoning.

%% file: sec/5_metrics.tex
\section{Metrics for Evidence Sensitivity}
\label{sec:metrics}
These metrics jointly characterize both prediction accuracy and the degree to which model outputs depend on evidence.

\noindent\textbf{Accuracy.}
We compute accuracy on a joined set $\mathcal{J}$, consisting of model outputs and matched ViewDiag GT, yielding paired $(\hat{y}_i, y_i)$ values. 
For predictions $\hat{y}_i$ and targets $y_i$ over the joined set $\mathcal{J}$, we report
\begin{equation}
\mathrm{MAE} = \frac{1}{|\mathcal{J}|}\sum_{i\in\mathcal{J}} |\hat{y}_i - y_i|.
\end{equation}
We also report an L2-calibrated variant using a fitted scalar $s^*$:
\begin{equation}
s^* = \frac{\sum_i \hat{y}_i y_i}{\sum_i \hat{y}_i^2}, \quad
\mathrm{MAE}_{\mathrm{L2}} = \frac{1}{|\mathcal{J}|}\sum_i |s^*\hat{y}_i - y_i|.
\end{equation}
This isolates global scale bias (e.g., consistent over/underestimation) so we can compare models on accuracy after a single scalar correction, while still reporting raw MAE to avoid overstating calibration gains.
The scale $s^*$ is fitted per model on the joined set as a post-hoc correction without updating model parameters. 
When uncertainty is reported, we use bootstrap 95\% confidence intervals~\cite{efron1994introduction}.

\noindent\textbf{Consistency via distribution collapse.}
We measure prediction concentration using top-1 mass and effective support. Let $p(y)$ denote the empirical prediction distribution. The top-1 mass is $\max_y p(y)$, and the effective support is
\begin{equation}
\mathrm{EffSupport} = \frac{1}{\sum_y p(y)^2}.
\end{equation}
High top-1 mass and low effective support indicate a collapsed output distribution, consistent with prior-dominated predictions that respond weakly to evidence.

\noindent\textbf{Collapse quadrant.}
We visualize accuracy and consistency in a collapse quadrant (Figure~\ref{fig:teaser}). Models with both high error and high concentration occupy a collapse regime indicative of evidence-insensitive behavior. The remaining quadrants capture trade-offs such as accurate-but-peaky or inaccurate-but-diverse predictions.

\begin{figure*}[!t]
	\centering
	\begin{minipage}{0.32\linewidth}
		\centering
		\includegraphics[width=\linewidth]{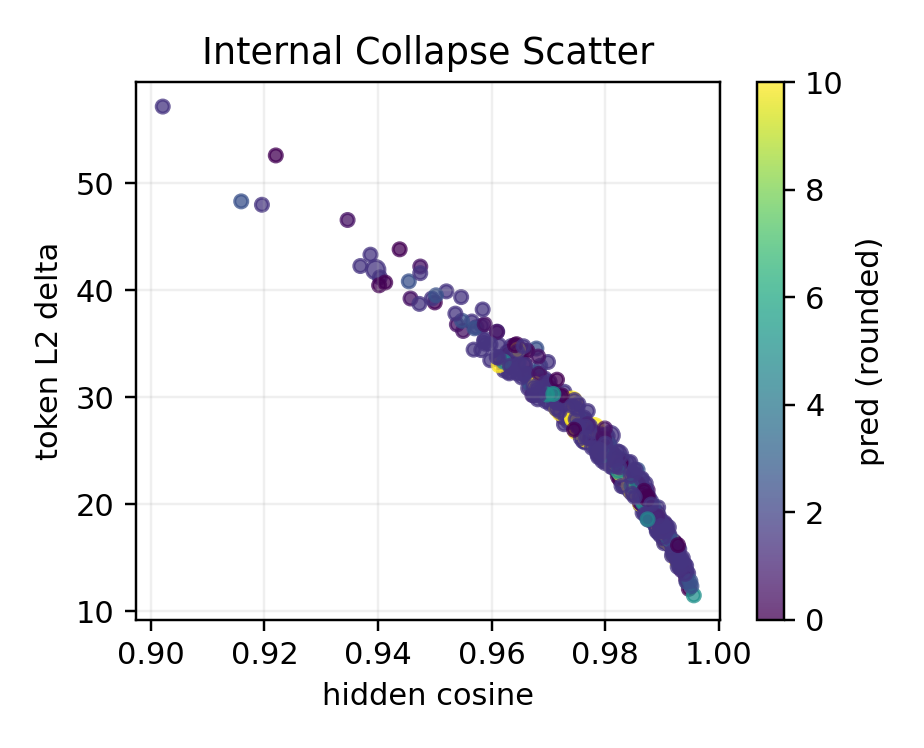}
		\small Qwen2-VL-7B
	\end{minipage}
	\begin{minipage}{0.32\linewidth}
		\centering
		\includegraphics[width=\linewidth]{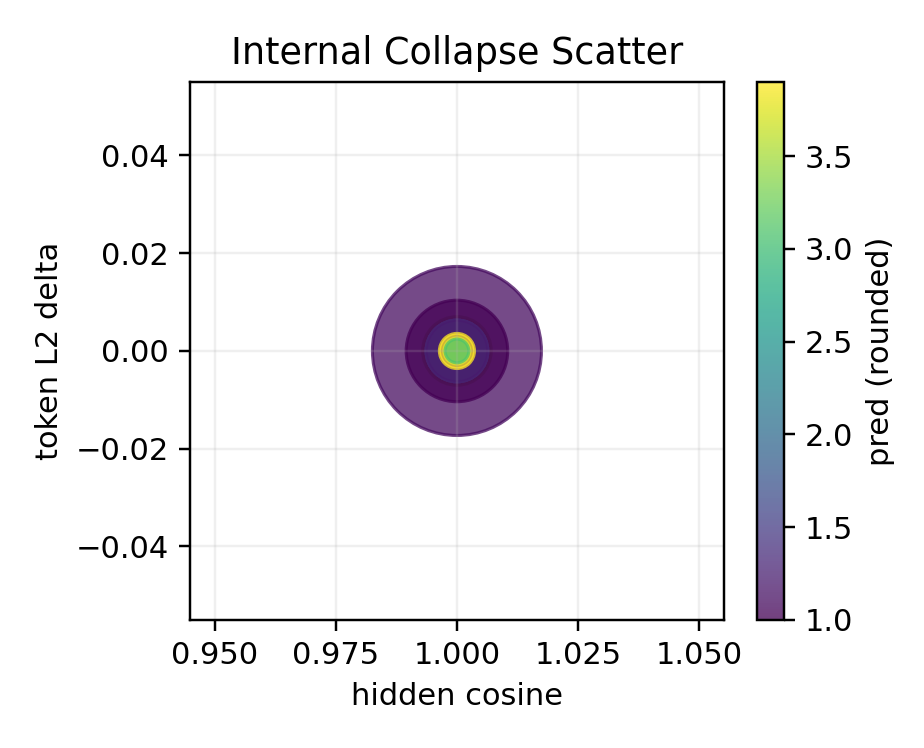}
		\small SpatialRGPT
	\end{minipage}
	\begin{minipage}{0.32\linewidth}
		\centering
		\includegraphics[width=\linewidth]{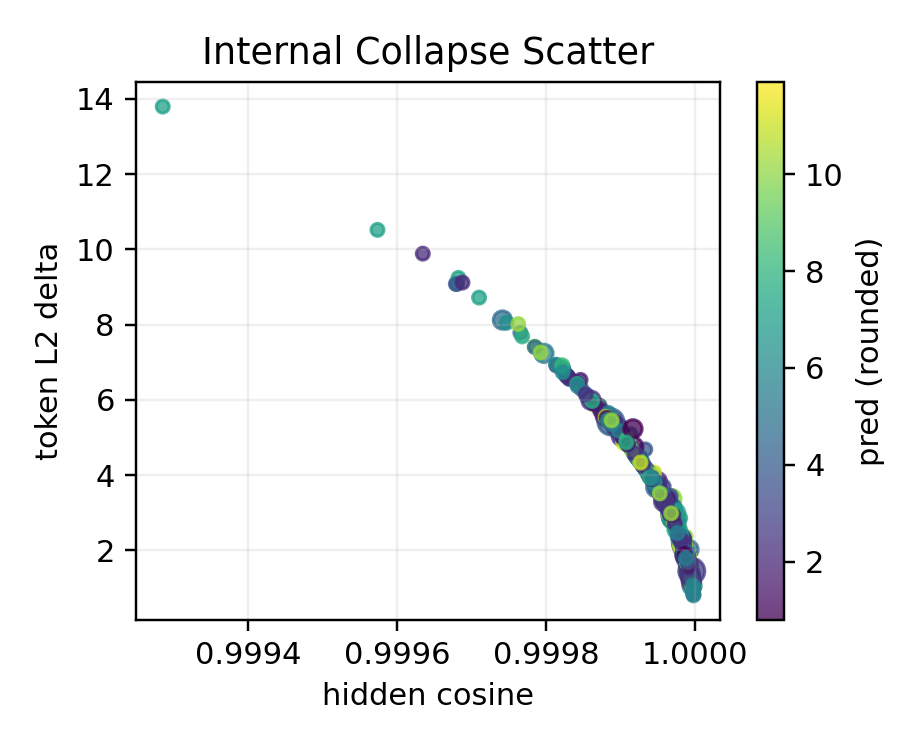}
		\small DepthLM
	\end{minipage}
	\caption{Internal collapse scatters for Qwen2-VL-7B, SpatialRGPT, and DepthLM ($n=512$ same-output pairs, bin size 0.1). Marker sizes reflect overlap counts. Qwen2-VL-7B and DepthLM exhibit decision collapse, while SpatialRGPT shows representation collapse (near-zero deltas).}
	\label{fig:internal_collapse_main}
\end{figure*}

\begin{figure*}[!t]
	\centering
	\begin{minipage}{0.32\linewidth}
		\centering
		\includegraphics[width=\linewidth,trim={0cm, 0cm, 0cm, 0.73cm}, clip]{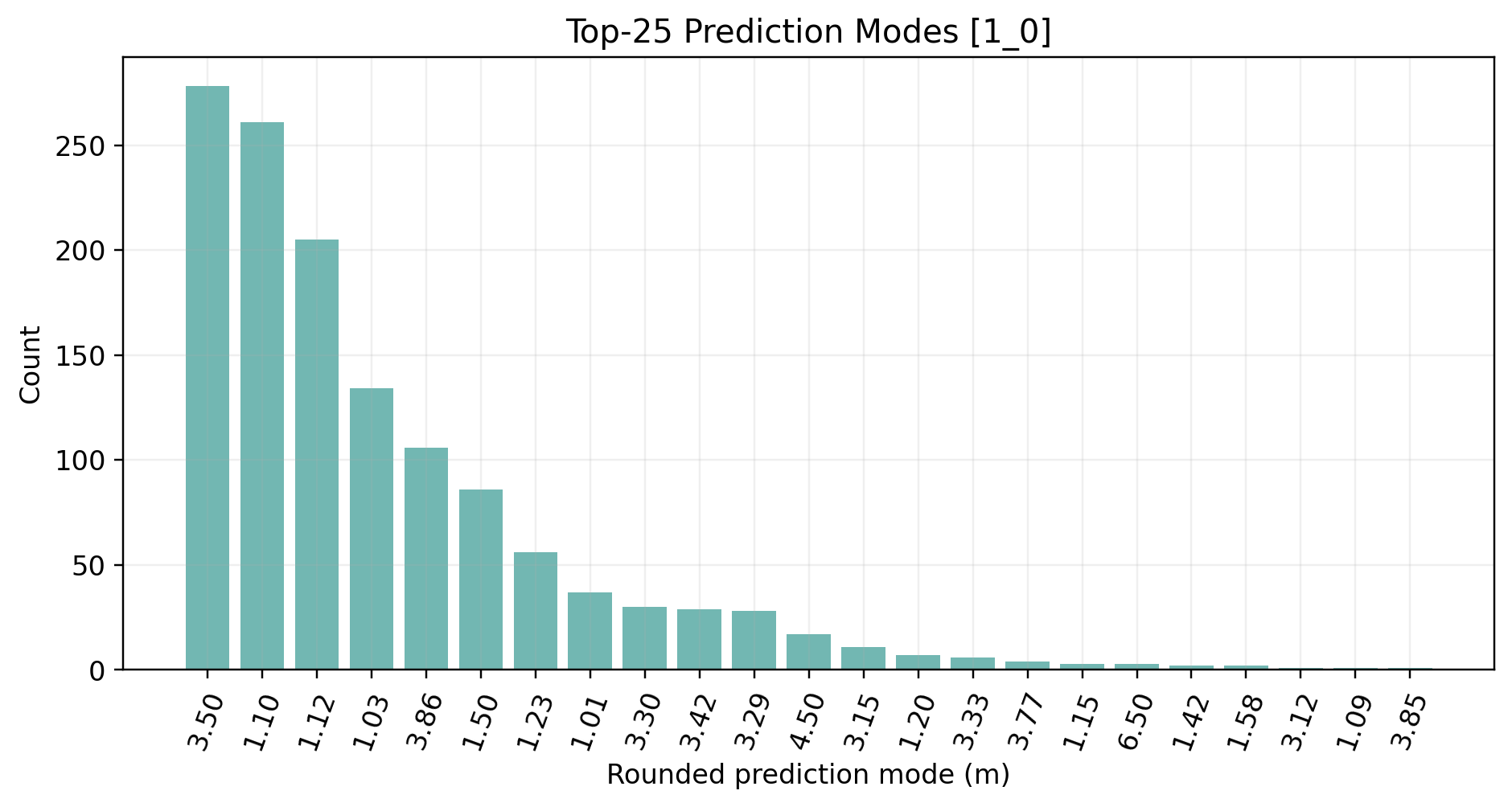}
		\small SpatialRGPT
	\end{minipage}
	\begin{minipage}{0.32\linewidth}
		\centering
		\includegraphics[width=\linewidth,trim={0cm, 0cm, 0cm, 0.73cm}, clip]{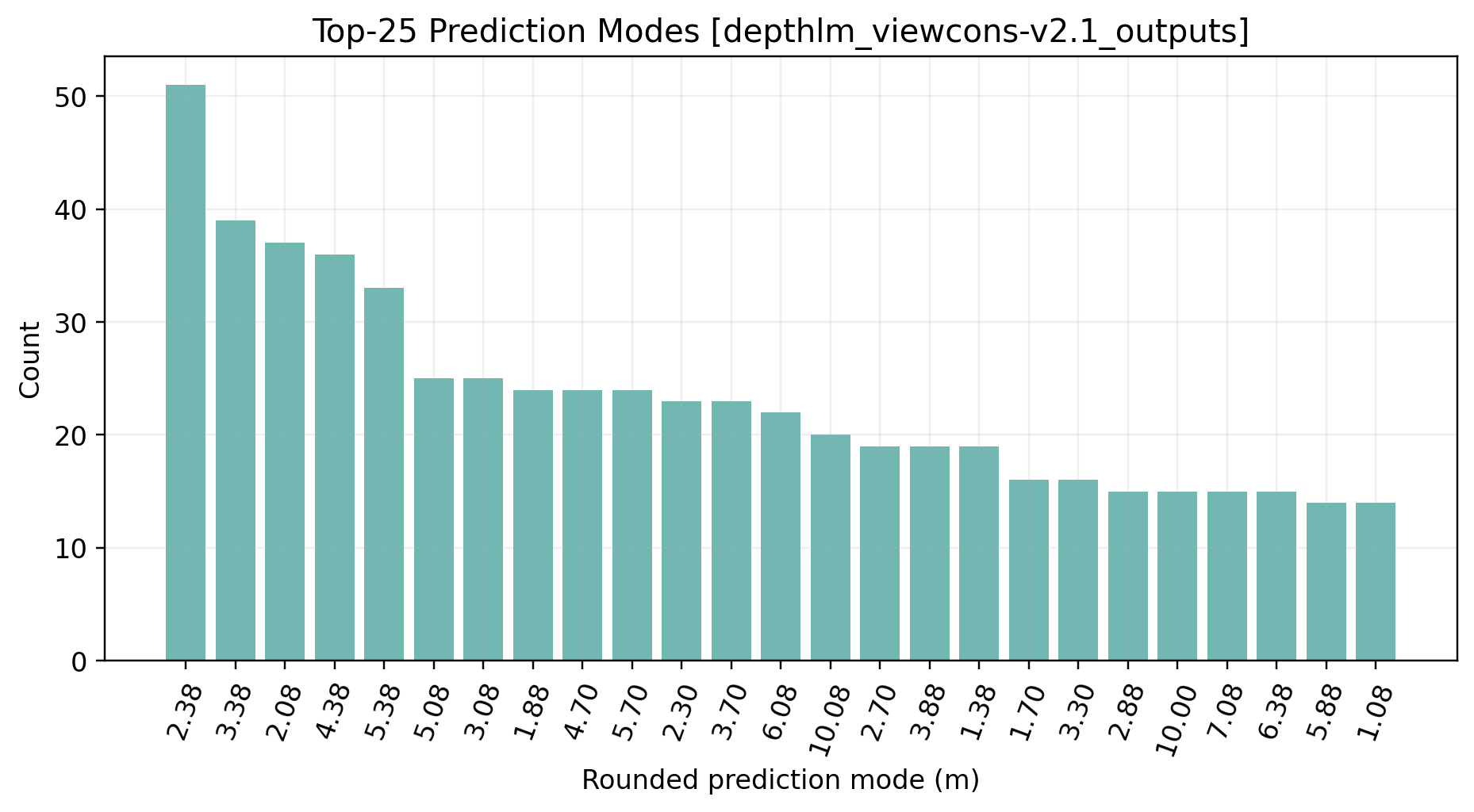}
		\small DepthLM
	\end{minipage}
	\begin{minipage}{0.32\linewidth}
		\centering
		\includegraphics[width=\linewidth,trim={0cm, 0cm, 0cm, 0.73cm}, clip]{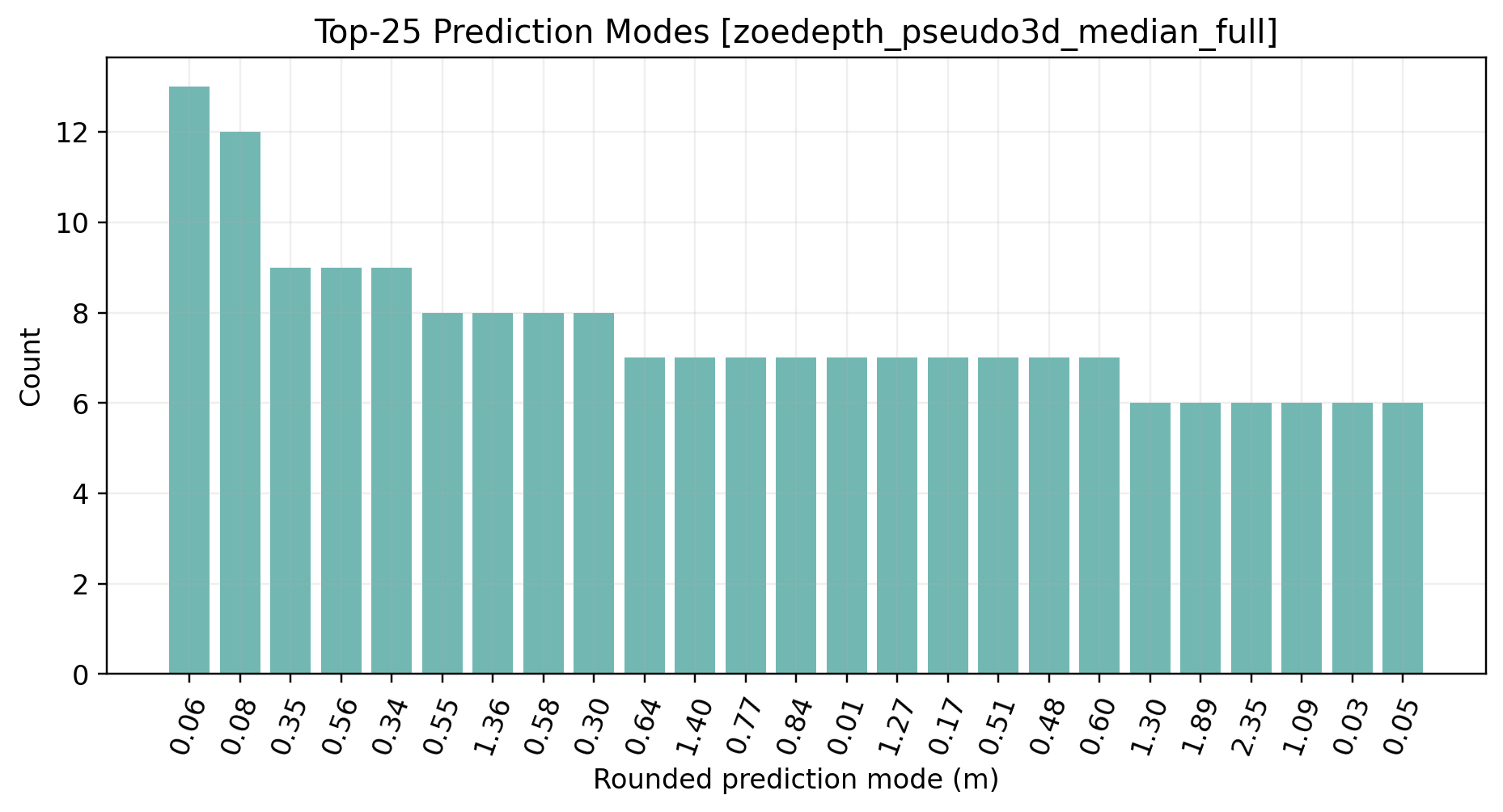}
		\small ZoeDepth
	\end{minipage}
	\caption{Top-25 mode profiles for SpatialRGPT, DepthLM, and ZoeDepth, highlighting output concentration in spatial VLMs.}
	\label{fig:mode_profile_models}
\end{figure*}

\begin{table}[!h]
\centering
\small
\begin{tabular}{lcc}
\toprule
Model & Dataset & MAE(raw) \\
\midrule
DepthLM~\cite{cai2025depthlm} & Hypersim & 3.14 \\
InternVL3-8B~\cite{internvl2024} & Hypersim & 3.35 \\
Qwen2-VL-7B~\cite{qwen2vl2024} & Hypersim & 3.65 \\
MiniCPM-V-4.5~\cite{minicpmv2024} & ScanNet & 0.58 \\
InternVL3-8B~\cite{internvl2024} & ScanNet & 0.71 \\
Qwen2-VL-7B~\cite{qwen2vl2024} & ScanNet & 1.03 \\
ZoeDepth~\cite{bhat2023zoedepth} & KITTI360 & 8.10 \\
DepthLM~\cite{cai2025depthlm} & KITTI360 & 15.34 \\
Qwen2-VL-7B~\cite{qwen2vl2024} & KITTI360 & 16.57 \\
\bottomrule
\end{tabular}
\caption{Per-dataset MAE for representative top models.}
\label{tab:perdataset}
\end{table}

\noindent\textbf{Evidence Sensitivity Score (ESS).}
We adopt a simple multiplicative proxy combining prediction error and output concentration.
Although heuristic, this score highlights regimes where both error and collapse are simultaneously high.
\begin{equation}
\mathrm{ESS} = \mathrm{MAE} \times \mathrm{Top1Mass}.
\end{equation}
Lower ESS indicates either accurate predictions or diverse, evidence-dependent outputs, while high ESS highlights collapse-prone behavior. We also report Top1Mass and EffSupport directly; the qualitative conclusions remain consistent across these measures.

\noindent\textbf{Internal collapse probe.}
We sample different-view pairs whose predicted distances match after binning (same-output; 0.1\,m bins by default) and compare their hidden states. Let $h_v$ be the pooled hidden vector for view $v$. We compute hidden cosine $\cos(h_{v_1},h_{v_2})$ and a token‑norm delta $\lVert h_{v_1}-h_{v_2}\rVert_2$. High cosine with near‑zero delta indicates a \emph{nearly invariant representation}, while high cosine with large deltas indicates \emph{decision collapse}. 

%% file: sec/5_experiments.tex
\section{Experimental Setup}
\label{sec:experiments}

This section summarizes the evaluated models, query formulation, and evaluation protocol.

\noindent\textbf{Models.}
We evaluate a diverse set of vision-language models spanning general-purpose and spatially specialized architectures: Qwen2-VL (2B/7B)~\cite{qwen2vl2024}, InternVL3 (1B/8B)~\cite{internvl2024}, MiniCPM-V (4.5/2.6)~\cite{minicpmv2024}, LLaVA (1.5/1.6)~\cite{liu2023llava}, VILA1.5-3B~\cite{lin2024vila}, SpatialRGPT~\cite{cheng2024spatialrgpt}, and DepthLM~\cite{cai2025depthlm}, which adapts VLMs for metric depth prediction. Main tables report the strongest variant per family, while smaller variants appear in internal probes and supplementary analyses. As a geometry-only baseline, we use ZoeDepth~\cite{bhat2023zoedepth}; unless noted, this refers to our pseudo-3D + median variant which uses the median predicted depth per region and unprojects the region centers to compute a pseudo-3D distance.

\noindent\textbf{Task and prompt format.}
The task asks the model to estimate the metric distance between two queried image regions. We enforce a consistent meter-valued answer format and normalize outputs into a unified numeric schema prior to scoring. 

\noindent\textbf{Inference settings.}
All experiments use deterministic decoding with a fixed prompt to isolate evidence sensitivity from sampling noise. Results are computed on aligned and valid pairs obtained by joining model outputs with ViewDiag metadata.

\noindent\textbf{Metrics.}
We report MAE (raw), L2-calibrated MAE, bootstrap confidence intervals, and distribution-collapse statistics (Top1Mass and EffSupport) as defined in Section~\ref{sec:metrics}.

%% file: sec/6_results.tex
\section{Results}
\label{sec:results}

We first examine qualitative evidence traces to illustrate the failure mode and then analyze aggregate statistics across the full benchmark.

\begin{figure}[t]
	\centering
	\includegraphics[width=\linewidth]{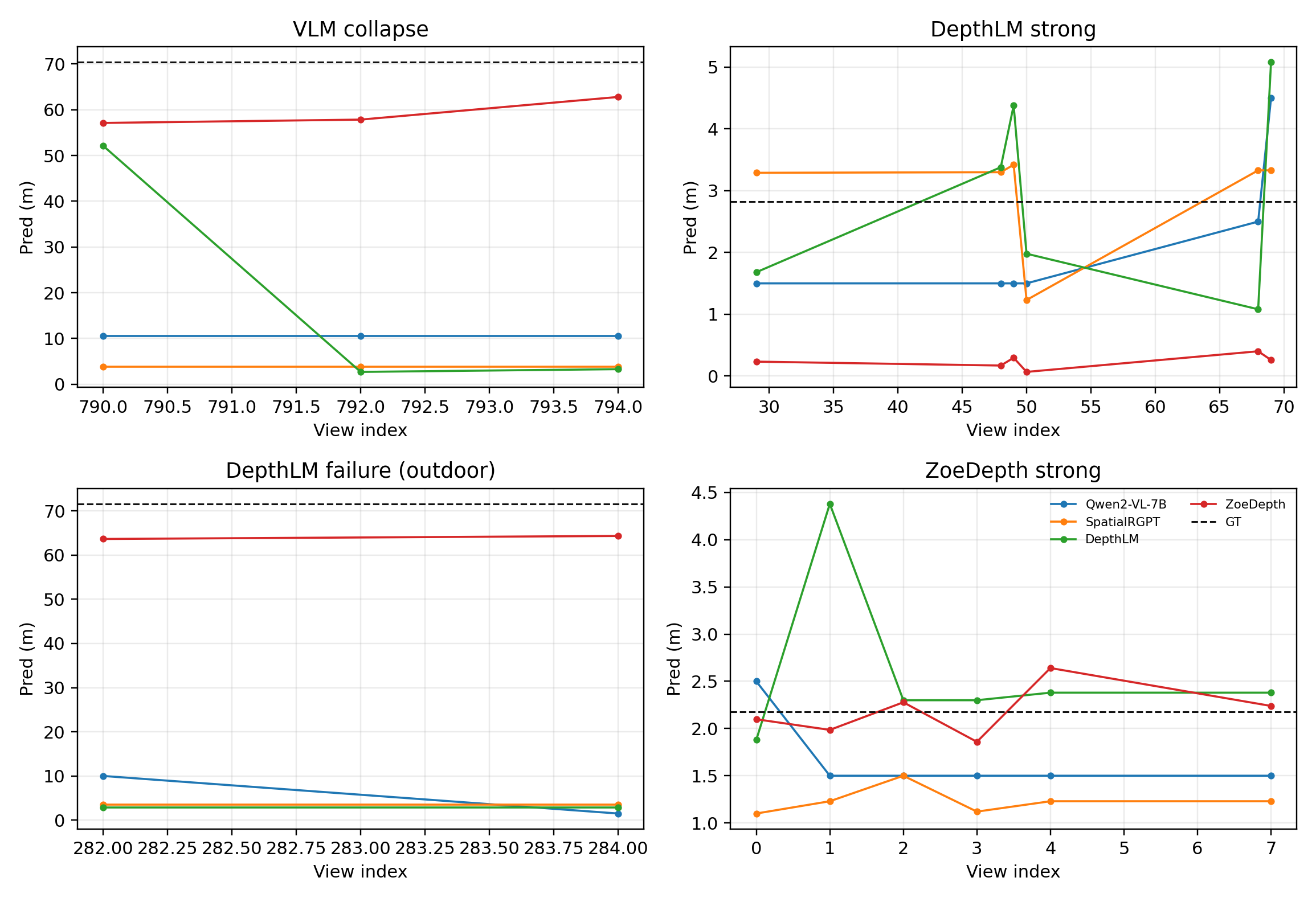}
	\caption{Targeted evidence traces: a VLM collapse case, a DepthLM strong case, a DepthLM failure case (outdoor), and a ZoeDepth success case. 
    }
	\label{fig:evidence_trace_grid}
\end{figure}

\noindent\textbf{Same-pair evidence traces.}
Figure~\ref{fig:teaser} contrasts indoor and outdoor examples, while Figure~\ref{fig:evidence_trace_grid} highlights cases that are representative but not exhaustive: VLM collapse, DepthLM success, DepthLM failure, and ZoeDepth success.  
\begin{table}[t]
\centering
\small
\begin{tabular}{lcc}
\toprule
Method & MAE(raw) & MAE(L2-calib) \\
\midrule
ZoeDepth~\cite{bhat2023zoedepth} & 4.59 & 4.48 \\
DepthLM~\cite{cai2025depthlm} & 5.77 & 5.83 \\
InternVL3-8B~\cite{internvl2024} & 6.22 & 6.51 \\
MiniCPM-V-4.5~\cite{minicpmv2024} & 6.40 & 6.47 \\
Qwen2-VL-7B~\cite{qwen2vl2024} & 6.42 & 5.89 \\
SpatialRGPT~\cite{cheng2024spatialrgpt} & 7.54 & 7.21 \\
LLaVA-1.6-Mistral-7B~\cite{liu2023llava} & 7.79 & 7.14 \\
MiniCPM-V-2.6~\cite{minicpmv2024} & 7.90 & 7.59 \\
VILA1.5-3B~\cite{lin2024vila} & 15.14 & 6.84 \\
\bottomrule
\end{tabular}
\caption{Full-set MAE for ViewDiag. MAE(raw) uses raw predictions; MAE(L2-calib) uses a fitted scalar. Lower is better.}
\label{tab:leaderboard}
\end{table}
These show a consistent pattern: geometry-based baselines respond to viewpoint changes, whereas VLM predictions often remain stable even when incorrect. DepthLM improves average accuracy but still fails on long-range outdoor pairs, indicating only partial evidence sensitivity. These examples are illustrative; all claims are supported by aggregate metrics and distributional analyses.

\begin{table*}[!t]
\centering
\small
\begin{tabular}{lccccc}
\toprule
Model & MAE(raw) & MAE raw CI95 & Top1Mass & EffSupport & ESS \\
\midrule
ZoeDepth~\cite{bhat2023zoedepth} & 4.59 & [4.07, 5.09] & 0.0099 & 449.79 & 0.0455 \\
DepthLM~\cite{cai2025depthlm} & 5.77 & [5.19, 6.38] & 0.0391 & 147.33 & 0.2256 \\
InternVL3-8B~\cite{internvl2024} & 6.22 & [5.57, 6.82] & 0.1942 & 12.16 & 1.2078 \\
MiniCPM-V-4.5~\cite{minicpmv2024} & 6.40 & [5.74, 7.06] & 0.2974 & 7.50 & 1.9037 \\
Qwen2-VL-7B~\cite{qwen2vl2024} & 6.42 & [5.78, 7.02] & 0.4014 & 7.19 & 2.5765 \\
LLaVA-1.6-Mistral-7B~\cite{liu2023llava} & 7.79 & [7.11, 8.51] & 0.5551 & 3.21 & 4.3245 \\
\bottomrule
\end{tabular}
\caption{Full-set collapse descriptors with bootstrap 95\% CI for MAE(raw). Top1Mass and EffSupport quantify output concentration; ESS combines error and concentration.}
\label{tab:ci_collapse}
\end{table*}

\noindent\textbf{Overall accuracy and collapse quadrant.}
Table~\ref{tab:leaderboard} reports full-set MAE. Geometry baselines achieve the strongest raw accuracy, while InternVL3-8B performs best among the evaluated VLMs. The consistency–accuracy plot in the teaser (Figure~\ref{fig:teaser}) shows that several VLMs occupy the high-consistency, high-error region, indicating evidence-insensitive collapse rather than robust geometric grounding.

Consistency itself is not problematic: geometry baselines produce stable predictions when they are also accurate. In contrast, high consistency without accuracy, particularly under viewpoint changes, signals collapse.

\noindent\textbf{Distribution behavior and uncertainty.}
Prediction distributions reveal substantial output concentration in several VLMs despite competitive mean errors. For example, Qwen2-VL-7B shows top-1 mass 0.4014 with effective support 7.19, whereas InternVL3-8B is less concentrated (top-1 mass 0.1942, support 12.16). Bootstrap confidence intervals overlap among the strongest VLMs, suggesting that small MAE differences should not be over-interpreted. We therefore also report the Evidence Sensitivity Score (ESS), which combines error and concentration.

\noindent\textbf{Distributional collapse signatures.}
Figure~\ref{fig:teaser} illustrates a typical collapse pattern: biased ground-truth–prediction density combined with stable predictions across views. Figure~\ref{fig:residual_depth_baselines} compares residuals across SpatialRGPT, DepthLM, and ZoeDepth. Figures~\ref{fig:pred_hist_models} and~\ref{fig:mode_profile_models} further reveal output concentration patterns; additional diagnostics for Qwen models are provided in the supplementary.

\noindent\textbf{Internal collapse diagnosis.}
Internal probes reveal that identical outputs often arise from different hidden states. In Qwen2-VL-7B and DepthLM, most sampled pairs fall into the decision-collapse regime, indicating unstable decision mappings over changing representations. SpatialRGPT instead shows representation collapse, where hidden states remain nearly identical across views.  

\noindent At $n=512$ sampled pairs, Wilson 95\% confidence intervals are tight: decision collapse for Qwen2-VL-7B and DepthLM is $1.0$ (CI $[0.993,1.000]$), while SpatialRGPT exhibits representation collapse with the same confidence bounds. Figure~\ref{fig:internal_collapse_main} visualizes the diagnostic. Each point corresponds to a pair of views with identical predicted distance. The x axis shows hidden-state cosine similarity and the y axis token-norm change; the decision-collapse regime corresponds to high cosine with large L2 deltas.

\noindent\textbf{Per-dataset behavior.}
Table~\ref{tab:perdataset} shows substantial dataset-specific ranking shifts. Hypersim and ScanNet are comparatively forgiving for VLMs (single-digit MAE), while KITTI360 dominates long-range error budgets. This difference reflects domain and range shifts and highlights that apparent robustness on indoor data does not transfer cleanly to outdoor long-range scenes.

\noindent\textbf{Key findings.}
\begin{itemize}
	\item Geometry baselines achieve the strongest raw MAE, while leading VLMs narrow the gap but remain highly concentrated (Tables~\ref{tab:leaderboard},~\ref{tab:ci_collapse}).
    \item Evidence traces reveal a consistent-yet-wrong regime where predictions remain stable across views (Figure~\ref{fig:evidence_trace_grid}).
	\item Residual and distribution analyses show that long-range errors and mode collapse are most pronounced for spatial VLMs, whereas geometry baselines retain broader output distributions (Figures~\ref{fig:residual_depth_baselines},~\ref{fig:pred_hist_models},~\ref{fig:mode_profile_models}).
	\item Internal-collapse diagnostics distinguish model families: Qwen2-VL-7B~\cite{qwen2vl2024} and DepthLM~\cite{cai2025depthlm} show decision collapse, while SpatialRGPT~\cite{cheng2024spatialrgpt} shows representation collapse (Figure~\ref{fig:internal_collapse_main}).
    \item Rankings vary across datasets, indicating that indoor gains do not reliably transfer to outdoor long-range scenes (Table~\ref{tab:perdataset}).
\end{itemize}

%% file: sec/7_analysis.tex
\section{Analysis and Discussion}
\label{sec:analysis}
Consistency across views is often treated as a sign of robustness, but our results show that this can be misleading. High consistency may co-occur with substantial error and strong output concentration, indicating that stable predictions do not necessarily reflect correct or geometrically grounded reasoning. When predictions remain unchanged despite changes in visual evidence and are still incorrect, consistency becomes a failure signal rather than a strength. We refer to this behavior as \emph{evidence insensitivity}: predictions are driven by narrow priors rather than geometric cues.

A likely explanation lies in the training dynamics of large VLMs. Language supervision and dataset biases may encourage models to internalize stereotyped distance priors. When geometric cues are ambiguous, especially in long-range outdoor scenes, models may fall back to a small set of preferred outputs that remain stable across views, producing an appearance of robustness while masking systematic inaccuracy. Our internal-collapse probe supports this interpretation: for Qwen2-VL-7B and DepthLM, identical outputs often arise from noticeably different hidden states, suggesting many-to-one decision mappings rather than stable geometric representations.

These findings have important implications for evaluation. Accuracy alone is insufficient, since a model may appear stable across viewpoints while remaining systematically wrong. Hybrid approaches such as DepthLM reduce average error, indicating that metric supervision helps, but do not eliminate evidence-insensitive collapse. This suggests that geometric signals improve predictions without guaranteeing evidence-sensitive reasoning.

We therefore argue that spatial VLM evaluation should jointly measure \emph{accuracy} and \emph{evidence sensitivity}. In practice, this means complementing standard error metrics with distributional concentration measures and at least one probe that distinguishes genuine geometric invariance from collapse driven by output priors. Additional analyses supporting these conclusions are included in the Supplementary.

%% file: sec/8_limitations.tex
\section{Limitations and Future Work}
\label{sec:limitations}

Our contribution is diagnostic rather than corrective: we characterize evidence-insensitive behavior but do not propose a mitigation method. The internal-collapse probe is limited to models with accessible hidden representations, and our evidence-sensitivity analysis relies on proxy metrics derived from prediction behavior. ViewDiag is also limited in scope, spanning three datasets and a single task, fixed-region distance estimation, which bounds our empirical claims. Still, the collapse signatures are consistent across indoor and outdoor settings as well as short- and long-range regimes.

Future work will extend these diagnostics to additional spatial reasoning tasks and incorporate pose-aware geometric perturbations to test evidence dependence under controlled view changes. Another promising direction is to develop training or inference interventions that encourage stronger evidence use while preserving accuracy.

%% file: sec/9_conclusion.tex
\section{Conclusion}
\label{sec:conclusion}
We study whether view consistency in spatial VLMs reflects genuine geometric grounding and find that it often signals evidence-insensitive collapse instead. Across model families, predictions frequently remain stable across viewpoints despite substantial error, placing many models in a regime marked by high consistency, strong output concentration, and low accuracy.

ViewDiag and our diagnostic suite provide a controlled framework for exposing this failure mode by jointly measuring accuracy, distributional collapse, and internal response to viewpoint change. Our results suggest that robustness in spatial VLMs should be judged not by output stability alone, but by whether predictions remain both accurate and meaningfully coupled to visual evidence.

%% file: main.bib
@String(AAAI = {AAAI})

@inproceedings{radford2021clip,
  title={Learning transferable visual models from natural language supervision},
  author={Radford, Alec and Kim, Jong Wook and Hallacy, Chris and Ramesh, Aditya and Goh, Gabriel and Agarwal, Sandhini and Sastry, Girish and Askell, Amanda and Mishkin, Pamela and Clark, Jack and others},
  booktitle={International Conference on Machine Learning},
  pages={8748--8763},
  year={2021},
  organization={PmLR}
}

@inproceedings{li2022blip,
  title={Blip: Bootstrapping language-image pre-training for unified vision-language understanding and generation},
  author={Li, Junnan and Li, Dongxu and Xiong, Caiming and Hoi, Steven},
  booktitle={International Conference on Machine Learning},
  pages={12888--12900},
  year={2022},
  organization={PMLR}
}

@inproceedings{li2023blip2,
  title={Blip-2: Bootstrapping language-image pre-training with frozen image encoders and large language models},
  author={Li, Junnan and Li, Dongxu and Savarese, Silvio and Hoi, Steven},
  booktitle={International Conference on Machine Learning},
  pages={19730--19742},
  year={2023},
  organization={PMLR}
}

@article{alayrac2022flamingo,
  title={Flamingo: a visual language model for few-shot learning},
  author={Alayrac, Jean-Baptiste and Donahue, Jeff and Luc, Pauline and Miech, Antoine and Barr, Iain and Hasson, Yana and Lenc, Karel and Mensch, Arthur and Millican, Katie and Reynolds, Malcolm and others},
  journal={Advances in Neural Information Processing Systems},
  volume={35},
  pages={23716--23736},
  year={2022}
}

@inproceedings{chen2024internvl,
  title={Internvl: Scaling up vision foundation models and aligning for generic visual-linguistic tasks},
  author={Chen, Zhe and Wu, Jiannan and Wang, Wenhai and Su, Weijie and Chen, Guo and Xing, Sen and Zhong, Muyan and Zhang, Qinglong and Zhu, Xizhou and Lu, Lewei and others},
  booktitle={Proceedings of the IEEE/CVF Conference on Computer Vision and Pattern Recognition},
  pages={24185--24198},
  year={2024}
}

@article{hong2023d3llm,
  title={3d-llm: Injecting the 3d world into large language models},
  author={Hong, Yining and Zhen, Haoyu and Chen, Peihao and Zheng, Shuhong and Du, Yilun and Chen, Zhenfang and Gan, Chuang},
  journal={Advances in Neural Information Processing Systems},
  volume={36},
  pages={20482--20494},
  year={2023}
}

@inproceedings{johnson2017clevr,
  title={Clevr: A diagnostic dataset for compositional language and elementary visual reasoning},
  author={Johnson, Justin and Hariharan, Bharath and Van Der Maaten, Laurens and Fei-Fei, Li and Lawrence Zitnick, C and Girshick, Ross},
  booktitle={Proceedings of the IEEE/CVF Conference on Computer Vision and Pattern Recognition},
  pages={2901--2910},
  year={2017}
}

@inproceedings{hudson2019gqa,
  title={Gqa: A new dataset for real-world visual reasoning and compositional question answering},
  author={Hudson, Drew A and Manning, Christopher D},
  booktitle={Proceedings of the IEEE/CVF Conference on Computer Vision and Pattern Recognition},
  pages={6700--6709},
  year={2019}
}

@inproceedings{yeh2025allangles,
  title={Seeing from another perspective: Evaluating multi-view understanding in mllms},
  author={Yeh, Chun-Hsiao and Wang, Chenyu and Tong, Shengbang and Cheng, Ta-Ying and Wang, Ruoyu and Chu, Tianzhe and Zhai, Yuexiang and Chen, Yubei and Gao, Shenghua and Ma, Yi},
  booktitle={Proceedings of the AAAI Conference on Artificial Intelligence},
  volume={40},
  number={14},
  pages={12000--12008},
  year={2026}
}

@inproceedings{ranftl2021dpt,
  title={Vision transformers for dense prediction},
  author={Ranftl, Ren{\'e} and Bochkovskiy, Alexey and Koltun, Vladlen},
  booktitle={Proceedings of the IEEE/CVF International Conference on Computer Vision},
  pages={12179--12188},
  year={2021}
}

@inproceedings{yang2024depthanything,
  title={Depth anything: Unleashing the power of large-scale unlabeled data},
  author={Yang, Lihe and Kang, Bingyi and Huang, Zilong and Xu, Xiaogang and Feng, Jiashi and Zhao, Hengshuang},
  booktitle={Proceedings of the IEEE/CVF Conference on Computer Vision and Pattern Recognition},
  pages={10371--10381},
  year={2024}
}

@inproceedings{godard2019monodepth2,
  title={Digging into self-supervised monocular depth estimation},
  author={Godard, Cl{\'e}ment and Mac Aodha, Oisin and Firman, Michael and Brostow, Gabriel J},
  booktitle={Proceedings of the IEEE/CVF International Conference on Computer Vision},
  pages={3828--3838},
  year={2019}
}

@inproceedings{chen2024spatialvlm,
  title={Spatialvlm: Endowing vision-language models with spatial reasoning capabilities},
  author={Chen, Boyuan and Xu, Zhuo and Kirmani, Sean and Ichter, Brain and Sadigh, Dorsa and Guibas, Leonidas and Xia, Fei},
  booktitle={Proceedings of the IEEE/CVF Conference on Computer Vision and Pattern Recognition},
  pages={14455--14465},
  year={2024}
}

@article{cheng2024spatialrgpt,
  title={Spatialrgpt: Grounded spatial reasoning in vision-language models},
  author={Cheng, An-Chieh and Yin, Hongxu and Fu, Yang and Guo, Qiushan and Yang, Ruihan and Kautz, Jan and Wang, Xiaolong and Liu, Sifei},
  journal={Advances in Neural Information Processing Systems},
  volume={37},
  pages={135062--135093},
  year={2024}
}

@inproceedings{wu2025indoor,
  title={From indoor to open world: Revealing the spatial reasoning gap in mllms},
  author={Wu, Mingrui and Wang, Zhaozhi and Wang, Fangjinhua and Yang, Jiaolong and Pollefeys, Marc and Zhang, Tong},
  booktitle={Proceedings of the IEEE/CVF Conference on Computer Vision and Pattern Recognition},
  pages={16789--16799},
  year={2026}
}

@article{feng2025seeing,
  title={Seeing across views: Benchmarking spatial reasoning of vision-language models in robotic scenes},
  author={Feng, Zhiyuan and Kang, Zhaolu and Wang, Qijie and Du, Zhiying and Yan, Jiongrui and Shi, Shubin and Yuan, Chengbo and Liang, Huizhi and Deng, Yu and Li, Qixiu and others},
  journal={arXiv preprint arXiv:2510.19400},
  year={2025}
}

@inproceedings{li2018megadepth,
  title={Megadepth: Learning single-view depth prediction from internet photos},
  author={Li, Zhengqi and Snavely, Noah},
  booktitle={Proceedings of the IEEE/CVF Conference on Computer Vision and Pattern Recognition},
  pages={2041--2050},
  year={2018}
}

@article{bhat2023zoedepth,
  title={Zoedepth: Zero-shot transfer by combining relative and metric depth},
  author={Bhat, Shariq Farooq and Birkl, Reiner and Wofk, Diana and Wonka, Peter and M{\"u}ller, Matthias},
  journal={arXiv preprint arXiv:2302.12288},
  year={2023}
}

@book{efron1994introduction,
  title={An introduction to the bootstrap},
  author={Efron, Bradley and Tibshirani, Robert J},
  year={1993},
  publisher={Chapman and Hall/CRC}
}

@article{liu2023llava,
  title={Visual instruction tuning},
  author={Liu, Haotian and Li, Chunyuan and Wu, Qingyang and Lee, Yong Jae},
  journal={Advances in Neural Information Processing Systems},
  volume={36},
  pages={34892--34916},
  year={2023}
}

@article{qwen2vl2024,
  title={Qwen2-vl: Enhancing vision-language model's perception of the world at any resolution},
  author={Wang, Peng and Bai, Shuai and Tan, Sinan and Wang, Shijie and Fan, Zhihao and Bai, Jinze and Chen, Keqin and Liu, Xuejing and Wang, Jialin and Ge, Wenbin and others},
  journal={arXiv preprint arXiv:2409.12191},
  year={2024}
}

@article{internvl2024,
  title={Internvl3: Exploring advanced training and test-time recipes for open-source multimodal models},
  author={Zhu, Jinguo and Wang, Weiyun and Chen, Zhe and Liu, Zhaoyang and Ye, Shenglong and Gu, Lixin and Tian, Hao and Duan, Yuchen and Su, Weijie and Shao, Jie and others},
  journal={arXiv preprint arXiv:2504.10479},
  year={2025}
}

@article{minicpmv2024,
  title={Minicpm-v: A gpt-4v level mllm on your phone},
  author={Yao, Yuan and Yu, Tianyu and Zhang, Ao and Wang, Chongyi and Cui, Junbo and Zhu, Hongji and Cai, Tianchi and Li, Haoyu and Zhao, Weilin and He, Zhihui and others},
  journal={arXiv preprint arXiv:2408.01800},
  year={2024}
}

@inproceedings{lin2024vila,
  title={Vila: On pre-training for visual language models},
  author={Lin, Ji and Yin, Hongxu and Ping, Wei and Molchanov, Pavlo and Shoeybi, Mohammad and Han, Song},
  booktitle={Proceedings of the IEEE/CVF Conference on Computer Vision and Pattern Recognition},
  pages={26689--26699},
  year={2024}
}

@inproceedings{roberts2021hypersim,
  title={Hypersim: A photorealistic synthetic dataset for holistic indoor scene understanding},
  author={Roberts, Mike and Ramapuram, Jason and Ranjan, Anurag and Kumar, Atulit and Bautista, Miguel Angel and Paczan, Nathan and Webb, Russ and Susskind, Joshua M},
  booktitle={Proceedings of the IEEE/CVF International Conference on Computer Vision},
  pages={10912--10922},
  year={2021}
}

@inproceedings{dai2017scannet,
  title={Scannet: Richly-annotated 3d reconstructions of indoor scenes},
  author={Dai, Angela and Chang, Angel X and Savva, Manolis and Halber, Maciej and Funkhouser, Thomas and Nie{\ss}ner, Matthias},
  booktitle={Proceedings of the IEEE/CVF Conference on Computer Vision and Pattern Recognition},
  pages={5828--5839},
  year={2017}
}

@article{liao2021kitti360,
  title={Kitti-360: A novel dataset and benchmarks for urban scene understanding in 2d and 3d},
  author={Liao, Yiyi and Xie, Jun and Geiger, Andreas},
  journal={IEEE Transactions on Pattern Analysis and Machine Intelligence},
  volume={45},
  number={3},
  pages={3292--3310},
  year={2023},
  publisher={IEEE}
}

@article{geirhos2020shortcut,
  title={Shortcut learning in deep neural networks},
  author={Geirhos, Robert and Jacobsen, J{\"o}rn-Henrik and Michaelis, Claudio and Zemel, Richard and Brendel, Wieland and Bethge, Matthias and Wichmann, Felix A},
  journal={Nature Machine Intelligence},
  volume={2},
  number={11},
  pages={665--673},
  year={2020},
  publisher={Nature Publishing Group UK London}
}

@article{cai2025depthlm,
  title={Depthlm: Metric depth from vision language models},
  author={Cai, Zhipeng and Yeh, Ching-Feng and Xu, Hu and Liu, Zhuang and Meyer, Gregory and Lei, Xinjie and Zhao, Changsheng and Li, Shang-Wen and Chandra, Vikas and Shi, Yangyang},
  journal={arXiv preprint arXiv:2509.25413},
  year={2025}
}
